%% file: ms.tex
\documentclass[conference]{IEEEtran}
\usepackage{times}

\usepackage[numbers]{natbib}
\usepackage{multicol}
\usepackage[bookmarks=true]{hyperref}
\usepackage{graphicx} 
\usepackage{epsfig} 
\usepackage{times} 
\usepackage{amsmath} 
\usepackage{amssymb}  
\usepackage{subcaption}
\usepackage[font={small}]{caption}
\usepackage{bm}
\usepackage{booktabs}

\newcommand{\D}{d}

\newcommand{\Oc}{\mathbf{o}}

\begin{document}

\title{MRFMap: Online Probabilistic 3D Mapping using Forward Ray Sensor Models}

\author{\authorblockN{Kumar Shaurya Shankar and Nathan Michael}
\authorblockA{The Robotics Institute\\
Carnegie Mellon University\\
Pittsburgh, PA 15213, USA\\
\texttt{\{kshaurya,nmichael\}@cmu.edu}}}

\maketitle

\begin{abstract}
    Traditional dense volumetric representations for robotic mapping make simplifying assumptions about sensor noise characteristics due to computational constraints.  We present a framework that, unlike conventional occupancy grid maps, explicitly models the sensor ray formation for a depth sensor via a Markov Random Field and performs loopy belief propagation to infer the marginal probability of occupancy at each voxel in a map. By explicitly reasoning about occlusions our approach models the correlations between adjacent voxels in the map. Further, by incorporating learnt sensor noise characteristics we perform accurate inference even with noisy sensor data without ad-hoc definitions of sensor uncertainty. We propose a new metric for evaluating probabilistic volumetric maps and demonstrate the higher fidelity of our approach on simulated as well as real-world datasets.
\end{abstract}

\IEEEpeerreviewmaketitle

\section{Introduction}
\input{intro.tex}

\section{Theory}
\input{theory.tex}

\section{Implementation}
\input{implementation.tex}

\section{Evaluation and Results}\label{sec:results}
\input{results.tex}

\section{Conclusion and Future Work}\label{sec:conclusion}
\input{conclusions.tex}

\section*{Acknowledgements}
The authors would like to thank Dr. Osman Ulusoy for their valuable insights and discussions.
\clearpage
\bibliographystyle{plainnat}
\bibliography{ms}

\input{supplement.tex}
\end{document}

%% file: intro.tex
Robotic navigation algorithms for real-world robots require accurate and dense probabilistic volumetric representations of the environment in order to plan and execute optimal traversal. Sensor data, however, always has associated acquisition noise and encoding this uncertainty within the map representation while still maintaining computational tractability is a key challenge in deploying these systems outside of controlled laboratory settings.

Occupancy grids are the conventional means of representing dense probabilistic volumetric data for robotic applications and are a discretised representation of binary labels probabilistically assigned to each cell denoting whether it is occupied or not~\cite{elfes1989using}. However, a fundamental assumption made for tractability of occupancy grids and its variants is the conditional independence of the occupancy of cells given sensor data. In the presence of sensor noise, or glancing rays, these methods end up blurring out the map details and clearing occupied space, respectively~\cite{hornung2013octomap}.

The seminal work of~\citet{thrun2003learning} argued that the occupancy mapping problem is essentially a high dimensional search in the space of all maps and should be tackled as such. By reasoning about the physics of sensor formation using forward models it is possible to reason in terms of the likelihood of the measurements for a given map hypothesis. The sensor data then directly drives a solution that explains the noisy observations as well as possible given the forward sensor model. However, this approach to mapping has historically been prohibitively expensive to compute in real-time.

\begin{figure}[ht]
    \centering
    \includegraphics[width=0.49\linewidth]{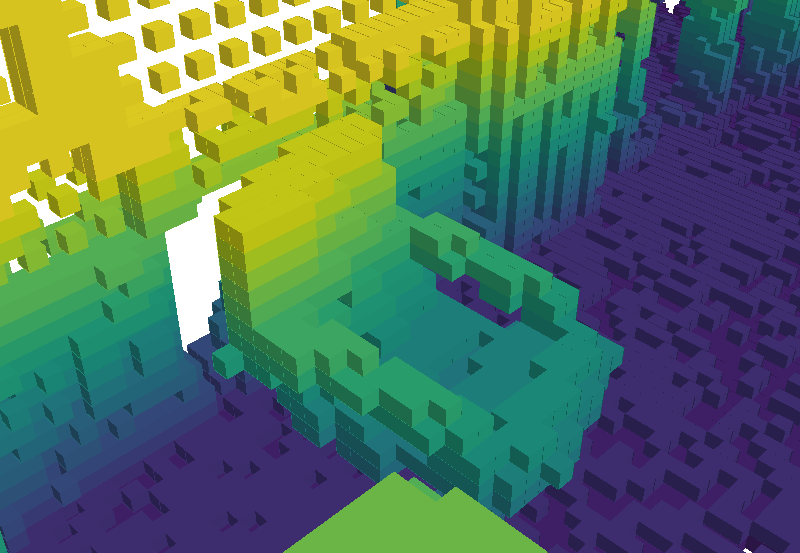}
    \includegraphics[width=0.49\linewidth]{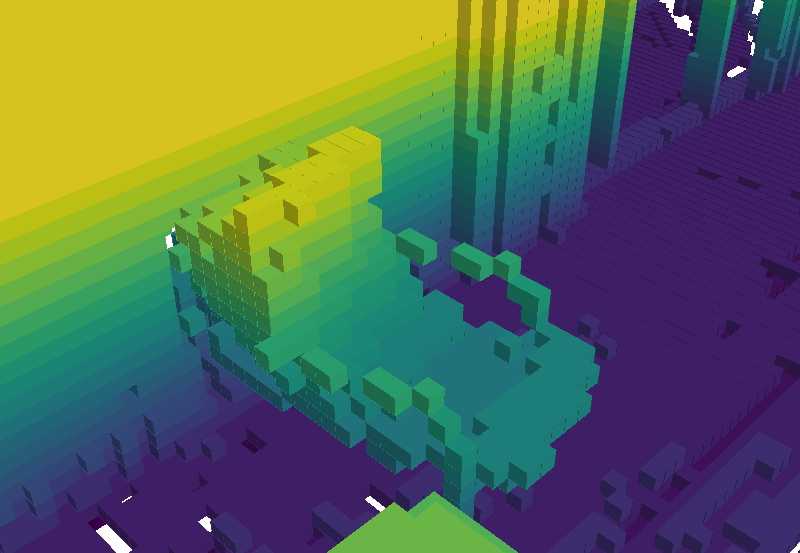}
        \caption{In the presence of sensor noise, or glancing rays, standard occupancy grid methods (OctoMap~\cite{hornung2013octomap}, Left) end up blurring out the map details and clearing occupied space, respectively. An MRFMap (Right) reasons about sensor ray formation and explicitly couples voxels both within each ray and between rays. Inset from Fig.~\ref{fig:ICL} from maps inferred using the augmented ICL-NUIM dataset~\cite{Choi_2015_CVPR} at \(0.05~\text{m}\) resolution. Occupied voxels coloured by height to highlight morphological differences.}\label{fig:teaser}
\end{figure}

We present a framework that explicitly reasons about the conditional dependence imposed on the occupancy of voxels traversed by each ray of a depth camera as a Markov Random Field (MRF). The tight intra-and inter-ray coupling explicitly incorporates conditional dependence of the occupancy of individual voxels as opposed to effectively marginalising out the sensor data in the form of log-odds updates as occupancy maps do. Visibility constraints imposed by using a forward sensor model enables simplification of the otherwise high dimensional inference. The forward model allows incorporating learnt sensor noise characteristics for more accurate inference. Finally, the inherent parallelisablity of both constructing the MRF and inferring it enable real-time performance on GPUs. 

Our key contributions are:
\begin{itemize}
    \item A Markov Random Field (MRF) based 3D occupancy grid framework that explicitly couples all voxels ray traced based on visibility;
    \item Incorporation of learnt sensor noise characteristics for improved map fidelity compared to occupancy grids;
    \item A principled evaluation metric for probabilistic occupancy maps; and
    \item An open-sourced\footnote{\url{https://mrfmap.github.io}} real-time GPU implementation.
\end{itemize}

\section{Related Work}
Although many representations exist for high fidelity surface mapping and SLAM, in this section we focus the discussion on volumetric probabilistic map representations that can be used directly within planning that explicitly represent free as well as occupied space.

Occupancy Grids~\cite{elfes1989using} are the canonical representation used for this purpose. Binary labelled cells denote occupied or unoccupied regions of the map. A Bayes filter is used to update the probability of occupancy for each cell independently. \citet{hornung2013octomap} introduced the OctoMap framework that uses octrees as an efficient data structure to implement variable resolution occupancy grids. However, by virtue of using very restrictive simplifying independence assumptions, measurement updates to neighbouring cells are not shared in occupancy grids, and adding noisy sensor data ends up blurring the map estimate. Variants such as Normal Distribution Transform Occupancy Maps~\cite{saarinen2013normal} store a Gaussian density within each cell to capture a more precise measure of the occupancy distribution. \citet{schulz2018efficient} present a real-time version of the same and add occlusion aware updates. Similarly, approaches exist that perform per cell filtering~\cite{saval2017review} that can be used to filter incoming sensor scan end points. Although the fidelity of the reconstruction in all these approaches is better than equivalent occupancy grid maps, they suffer the same drawbacks with independent cells not sharing information.

Gaussian Process Occupancy Maps~\cite{o2012gaussian} and their variants such as~\cite{wang2016fast,kim2012building} have been employed for estimating continuous occupancy maps by casting occupancy inference as a classification problem. By exploiting implicit structure in the world correlated sparse sensor measurements are used to reason about unobserved regions of space. Additionally, these maps can also encode sensor data and pose uncertainty. However, a major drawback of these approaches is the high computational complexity and memory usage that is prohibitive for real-time operation with dense 3D data, despite recent attempts to address the same~\cite{ramos2016hilbert,guo2019information}.
However, these approaches effectively marginalise out the sensor information to point samples of occupied and unoccupied space. Intuitively, reasoning about measurements as rays instead of endpoints provides more information that all such approaches discard.

Confidence-rich grid mapping~\cite{agha2019confidence} aims to enhance occupancy grids by encoding uncertainty within the grid and coupling voxels by reasoning about forward sensor models. The Bayes filter updates turn out to be identical to the ones used in~\cite{pollard2007change}, and thus share similar properties as with the MRFMap framework. The key difference between their approach and ours is that our framework allows for obtaining occupancy marginals based on poses being added or removed ad-hoc rather than relying on recursive Bayes filter updates, provides varying levels of resolution fidelity, and also permits the addition of visual data for enhancing inference, if required. Recently \citet{vespa2018efficient} introduced a novel hybrid representation that combines TSDFs and occupancy grids that although doesn't combine information from multiple rays explicitly, does use an implicit forward sensor model.

The theoretical foundations for performing inference using ray belief propagation trace their origins to methods that recover shape from silhouette cues~\cite{guan20083d, gargallo2007occupancy}. More recent work~\cite{liu2010ray, ulusoy2015towards, osman2016patches} use similar probabilistic foundations to come up with generative models of appearance and occupancy representations from multiple image views. However, for all these frameworks, depth (and thus occupancy) is a latent variable that is not directly observed. Our framework, although heavily inspired by~\citet{ulusoy2015towards, osman2016patches}, utilises depth images as opposed to using grayscale camera images. Since the variable of interest is directly observed, inference is much faster and possible to do in real-time. Further, we augment the formulation to incorporate learnt sensor noise to enable higher mapping fidelity.

%% file: theory.tex
We focus on sensors whose measurements can be modelled as ray measurements that return a single distance. For such sensors the uncertainty in angle is negligible in comparison to the uncertainty in distance, permitting the ray approximation instead of volumetric tracing. We also assume independence of rays in order to handle the updates separately for each ray.

We first present a brief introduction of depth ray potentials followed by the inference procedure, and finally the extension to incorporate learnt forward sensor model uncertainty.
\begin{figure}[ht]
	\centering
	\includegraphics[trim={0 0 0 0},clip,width=0.8\linewidth]{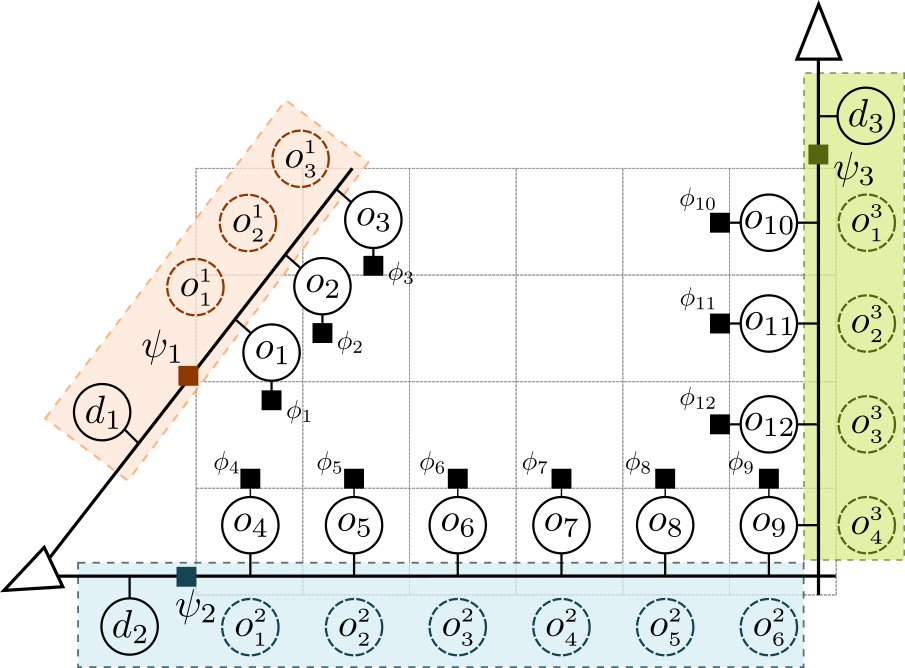}
	\caption{Graphical model illustration of the MRFMap framework. Each ray traversing through the voxels \(o_i\) establishes a depth potential \(\psi_r\) that connects all the occupancy nodes \(o_i^r\) it traverses through and an auxiliary depth variable \(\D_r\). Each voxel has a prior depth potential \(\phi_i\). Voxels connected by multiple rays (such as \(o_9\)) share information across the graph. Voxels within a shaded region are the \(\Oc_r\) voxels for that particular ray.}\label{fig:graph}
\end{figure}

\subsection{Ray Markov Random Field}
A Markov Random Field (MRF) is essentially a bipartite graph containing nodes corresponding to variables and factors that are connected by edges. Consider the ray MRF formation process as shown in Fig.~\ref{fig:graph}. Each ray from the camera sensor generates a ray potential that associates all the cells traversed by the ray. The cells are signified as nodes that can have a binary label \(o_i \in \{0,1\} \) signifying whether they are unoccupied or occupied, respectively. Similar to~\citet{ulusoy2015towards}, we specify a variable \(\D_r\) associated with the ray to represent the event that the depth of the first occupied cell along the direction of the ray is \(d_i^r\) (for \( o_i^r \), \( i \in 1 \dots N_r \) in increasing distance along a ray). However, we do not model the appearance of the voxels along the ray, since unlike their work, we have direct access to depth information. The joint distribution of this field is
\begin{equation}
	p( \Oc, \mathbf{\D} ) = \frac{1}{Z} \prod_{i \in \mathcal{X}} \phi_i(o_i) \prod_{r \in \mathcal{R}} \psi_r(\Oc_r, \D_r),
\end{equation}
where \(\mathcal{X}\) is the set of all the voxels \(o_i \in \{0,1\} \) corresponding to them being empty or occupied respectively, and \(\mathcal{R}\) is the set of all the rays from all the cameras viewing the scene, \(\Oc_r = \{o_1^r, \dots, o_{N_r}^r\} \) is the list of all the voxels traversed by a ray \(r\), \(\D_r\) is its corresponding depth variable, and \(Z\) is the normalisation constant. The total set of all the occupancy and depth variables are summarised as \(\Oc = \{o_i \mid i \in \mathcal{X}\} \) and \(\mathbf{d} = \{d_r \mid r \in \mathcal{R}\} \). \(\phi_i\) and \(\psi_r\) are the potential factors as described as follows:

\subsubsection{Prior Occupancy Factor}
This is simply a unary factor assigning an independent Bernoulli prior \(\gamma \) to the voxel occupancy label for each voxel
\begin{equation}
	\phi_i(o_i) = \gamma^{o_i}(1 - \gamma)^{1 - o_i}. \label{eq:prior_potential}
\end{equation}
In our experiments we use a non-informative prior however note that these per-voxel priors can be informed from per-pixel predictive methods if desired.

\subsubsection{Ray Depth Potential Factor}
A ray depth potential creates a factor graph connecting the binary occupancy variables \(o_i^r\) for all the voxels traversed by a single ray by virtue of making a measurement. Each of these voxels has a corresponding distance from the camera origin, and thus the ray is defined to include a depth variable \(\D_r\) that represents the event that the measured depth is at distance \(d_{i}^r\). For this to happen all the preceding voxels ought to be empty, i.e., have the value 0, and the corresponding voxel needs to be occupied, i.e., have a value 1. This leads to the definition of the joint occupancy and depth potential factor
\begin{equation}
	\psi_r(\Oc_r,\D_r ) = \begin{cases}
		\nu_r(d_{i}^r) & \text{if } d_r = \sum_{i=1}^{N_r} o_i^r \prod_{j<i} (1 - o_j^r) d_i^r \\
		0              & \text{otherwise}
	\end{cases}.\label{eq:depth_potential}
\end{equation}
Here \(\nu_r(d_{i}^r)\) denotes the probability of observing the measured depth measurement \(\mathcal{Z}_r\) if it originated from the latent depth variable \(d_{i}^r\) that we model as \begin{equation}
	\nu_r(d_{i}^r) = \mathcal{N}(\mathcal{Z}_r; d_{i}^r, \sigma(d_{i}^r)) . \label{eq:noise}
\end{equation}
This is our forward sensor model. This is based on the nature of RGB-D sensors, that are more precise compared to sonars that require different forward sensor models~\cite{thrun2003learning}. Similar to~\citet{basso2018robust} we model this probability to vary in mean and noise as a function of the distance from the camera, which enables utilising learnt sensor noise characteristics for most RGB-D sensors for accurate map inference.

This ray potential measures how well the occupancy and the depth variables explain the depth measurement \(\mathcal{Z}_r\). This is more apparent when Eq.~\ref{eq:depth_potential} is written as
\begin{equation} \label{eq:depth_potential_cases}
	\psi_r(\Oc_r,\D_r ) = \begin{cases}
		\nu_r(d_1^r) & \text{if}~ d^r = d_1^r, o_1^r = 1          \\
		\nu_r(d_2^r) & \text{if}~ d^r = d_2^r, o_1^r = 0, o_2 = 1 \\
		\vdots       &                                            \\
		\nu_r(d_N^r) & \text{if}~ d^r = d_N^r, o_1^r = 0,         \\&\dots, o_{{N_r}-1}^r = 0, o_{N_r}^r  = 1
	\end{cases}.
\end{equation}
This sparse structure of the ray potential enables simplification of the message passing equations by reducing the inference to a linear pass along the ray instead of exponential steps as detailed in the following subsections.

\subsection{Sum-Product Belief Propagation}
Sum-Product belief propagation~\cite{kschischang2001factor} is a common message-passing algorithm for performing approximate inference on cyclic factor graphs. By exploiting marginalisation of joint distributions using factorisation of a graph it enables computing marginal distributions very efficiently. Messages are passed back and forth between connected nodes and factors that try to influence the marginal belief of their neighbours. Although convergence is not guaranteed, in practice since the depth images are mostly consistent due to the relative accuracy of depth sensor measurements oscillations are rare and we obtain a solution quickly~\cite[p.~429]{koller2009probabilistic}.

The message sent from a variable node \(x\) to a factor \(f\) is the cumulative belief of all the incoming messages from factors to the node except the factor in question
\begin{equation}
	\mu_{x \rightarrow f}(x) = \prod_{g \in \mathcal{F}_x  \backslash f} \mu_{g \rightarrow x}(x) \label{eq:node_to_factor},
\end{equation}
where \(\mathcal{F}_x\) is the set of neighbouring factors to \(x\). Similarly, the message sent from the factor to the node is the marginalisation of the product of the value of the factor \(\phi_f\) with all the incoming messages from nodes other than the node in question
\begin{align}
	\mu_{f \rightarrow x}(x) = \sum_{\mathcal{X}_f \backslash x} \phi_f(\mathcal{X}_f) \prod_{y \in \mathcal{X}_f \backslash x} \mu_{y \rightarrow f}(y), \label{eq:factor_to_node}
\end{align}
where \(\mathcal{X}_f\) is the set of all neighbouring nodes of \(f\).

Upon convergence, the estimated marginal distribution of each node is proportional to the product of all messages from adjoining factors
\begin{equation}
	p(x) \propto \prod_{g \in \mathcal{F}_x} \mu_{g \rightarrow x}(x).
\end{equation}
Similarly, the joint marginal distribution of the set of nodes belonging to one factor is proportional to the product of the factor and the messages from the nodes
\begin{equation}
	p(\mathcal{X}_f) \propto \phi_f(\mathcal{X}_f) \prod_{\mathcal{X}_f} \mu_{x \rightarrow f}(x).
\end{equation}

\subsection{Messages to the variables}
\subsubsection{Ray Depth Potential to Depth Variable Messages}
Since we are only concerned about the depth variable, we marginalise out the messages from all the occupancy nodes going to the ray depth potential. Following Eq.~\ref{eq:factor_to_node} we have
\begin{equation}
	\mu_{\psi_r \rightarrow \D_r}\left(\D_r=d_{i}^r\right)=\sum_{o_{1}^r} \dots \sum_{ o_{N_r}^r} \psi_r\left(\mathbf{o}_r, \D_r\right) \prod_{j=1}^{N_r} \mu_{o_j^r \to \psi_r}\left(o_{j}^r\right).
\end{equation}
Na\"ively evaluating this equation would involve \(2^{N_r}\) evaluations for each ray (corresponding to the two possible event states for each \( o_j^r\)). However, we can exploit the sparse nature of the ray depth potential to recursively simplify this expression. After dropping the ray index for notational convenience and abbreviating \( \mu_{o_j^r \to \psi_r}(o_{j}^r)\) as \(\mu(o_j)\) we obtain (see the supplementary document~\cite{shankar20supp} for a detailed derivation)
\begin{align}
	\mu_{\psi \rightarrow \D}\left(\D=d_i\right)
	 & =  \nu{(d_i)} \mu(o_i=1)\prod_{k<i} \mu(o_k=0). \label{eq:depth_message}
\end{align}

\subsubsection{Depth Variable to Ray Depth Potential Messages}
Since the only factor connected to the depth variable is the ray depth potential itself, the message \( \mu_{\D_r \rightarrow \psi_r} \) is uninformative and, following Eq.~\ref{eq:node_to_factor}, it is set to a uniform value.

\subsubsection{Ray Depth Potential to Occupancy Variable Messages}
Similar to the depth variable messages, we marginalise out all the variables except the occupancy node in question
\begin{multline}
	\mu_{\psi_r \rightarrow o_{i}^r}\left(o_{i}^r=1\right)= \\
	\sum_{d_r} \sum_{ o_j^r \atop j \neq i}  \mu_{d_r\to\psi_r}(d_r) \psi_r\left(\mathbf{o}_r, \D_r\right) \prod_{j=1}^{N_r} \mu_{o_j^r \to \psi_r}\left(o_{j}^r\right).
\end{multline}

After simplification using the sparse structure of Eq.~\ref{eq:depth_potential} and dropping the ray indices for convenience we obtain (see the supplementary document~\cite{shankar20supp} for a detailed derivation)
\begin{align}
	\mu_{\psi \rightarrow o_{i}}\left(o_{i}=1\right) & =\sum_{j=1}^{i-1} \mu\left(o_{j}=1\right) \nu(d_j) \prod_{k<j} \mu\left(o_{k}=0\right) + \nonumber \\
	                                                 & \nu(d_i) \prod_{k<i} \mu\left(o_{k}=0\right) \label{eq:pos}
\end{align}
and
\begin{align}
	\mu_{\psi \rightarrow o_{i}}\left(o_{i}=0\right) & = \sum_{j=1}^{i-1} \mu\left(o_{j}=1\right) \nu(d_j) \prod_{k<j} \mu\left(o_{k}=0\right)   + \nonumber                     \\
	                                                 & \sum_{j=i+1}^{N} \frac{ \mu\left(o_{j}=1\right)}{\mu(o_i=0)} \nu(d_j) \prod_{k<j} \mu\left(o_{k}=0\right). \label{eq:neg}
\end{align}

To provide an intuition of what these messages do, the positive outgoing message in Eq.~\ref{eq:pos} sends a high value only if the likelihood of all the preceding voxels being empty is high (second term) after taking into account the possibility of each preceding voxel to be occupied (first term). Similarly, the negative outgoing message in Eq.~\ref{eq:neg} sends a high value if any of the subsequent voxels have a high likelihood of being occupied (second term) after taking into account the possibility of each preceding voxel to be occupied (first term). The first terms in both these equations are significant since they encapsulate the concept of visibility, i.e., as we traverse the ray through regions of occlusion, we can be less certain of the information the sensor measurement provides, and eventually the outgoing messages send out uninformative uniform messages.

\subsubsection{Occupancy Variable to Ray Depth Potential Messages}
Since other rays can (and often do) pass through the same occupancy variable node, the outgoing message \( \mu_{o_{i}^r \rightarrow \psi_r } \) is computed as per Eq.~\ref{eq:node_to_factor}. However, if a voxel is only traversed by a single ray, the only other factor that it receives messages from is the prior factor,~\( \phi_i \), described in Eq.~\ref{eq:prior_potential}.

\subsection{Evaluation Metric for Probabilistic Occupancy Maps}\label{sec:metric}
\begin{figure}[ht]
	\begin{center}
		\includegraphics[width=\linewidth]{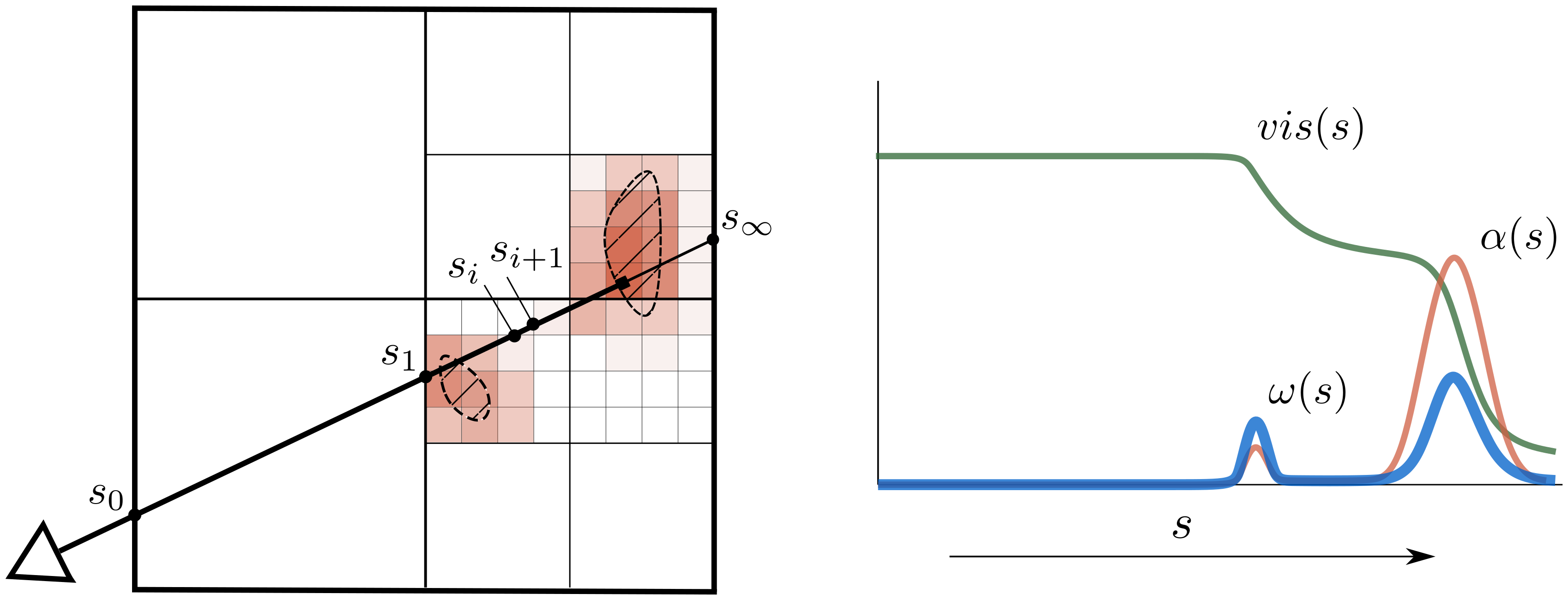}
		\caption{Piecewise continuous ray generating surface likelihood evaluation. For a camera ray parameterised by \(s\), the probability of occupancy per voxel can be considered a region with a constant occlusion density \(\alpha(s) \). The visibility decreases corresponding to the magnitude of \(\alpha(s) \) and the length traversed within. The probability of an occluding surface existing at a distance \(s\) is \(\omega(s)\), which is the product of the visibility and the map occlusion probability along the ray. Here, even though a voxel in the second region of the probabilistic map has higher probability mass of occupancy, it is obscured by a traversed region before it that reduces the likelihood of it being the generating surface for the ray depth measurement.}\label{fig:continuous}
	\end{center}
\end{figure}
In order to determine the accuracy of a given probabilistic occupancy map, especially one that has been generated from noisy data, the best map hypothesis should be one that maximises the likelihood of the sensor data. Further, it is not simply enough to look at thresholded occupancy values of a map and determine if sensor ray endpoints lie within an occupied voxel, as done in~\cite{hornung2013octomap}; one has to reason about visibility and occlusions by partially occupied voxels along the way. To approximate this,~\citet{stoyanov2013comparative} sampled negative free space to determine the fidelity of the reconstructed map. However, this is an ad-hoc approach and needs heuristics to determine what the region of the negative sampled distance along a ray should be. \citet{pathak20073d} introduced a visibility based measure that reasoned about occlusion. However their model does not take into account the fact that rays traverse through unequal lengths within a voxel, allowing the choice of discretisation to significantly influence the likelihood. To fix this, a continuous version of the same is presented by~\citet{crispell2010continuous} that models voxels as piecewise constant regions of occlusion density, which is what we present here in the context of probabilistic map evaluation.

As an analogue for the event that an occluding surface occurs at cell depth \(d_i\), \(\omega(s)\) is defined as the probability density function of the measurement generating surface along the ray length parameterised by a distance variable \(s\). The probabilistic volume is reimagined as a region of occlusion density, \(\alpha(s)\), that monotonically reduces the chances of the originating surface being at distance \(s\) as we move further along the ray (see Fig.~\ref{fig:continuous}).
The visibility at a distance \(s\) is defined as
\begin{equation}
	\operatorname{vis}(s)=e^{-\int_0^s \alpha(s') ds'}=e^{-\sum_{i=0}^{n-1} \alpha_{i} \ell_{i}+\alpha_{n}\left(s-s_{n}\right)},
\end{equation}
where \(\alpha_i\) denote the piecewise constant occlusion density, \(s_n\) denotes the distance of the ray upon hitting the \(n^{\text{th}}\) voxel, and \(l_i = s_{i+1} - s_i\) denote the lengths travelled within each voxel discretisation. This permits using voxels of differing dimensions while evaluating the accuracy.
The difference between two consecutive visibility probabilities across a voxel boundary then is equivalent to an occlusion probability assigned to it
\begin{equation}
	\Omega_{i}=\mathrm{vis}_{i}-\mathrm{vis}_{i+1}.
\end{equation}
The likelihood of the originating surface being at a particular distance \(s\) is proportional to the rate of change in the visibility along the ray as it traverses a region. This leads to the definition
\begin{align}
	\omega(s) & = \frac{d}{ds} (1 - \operatorname{vis}(s))            \\
	          & = \alpha(s) e^{-\int_0^s \alpha(s') ds'}    \nonumber \\
	\omega(s) & = \alpha(s) \operatorname{vis}(s).
\end{align}
Intuitively, even if there exist voxels along a ray with a higher probability of occupancy as per the map, the fact that the ray has already traversed through partially occluded regions previously reduce the probability that the measurement generating surface exists further along the ray (see Fig.~\ref{fig:continuous}). Note the similarity with the first terms in Eq.~\ref{eq:pos} and Eq.\ref{eq:neg}.

We thus define a ray depth measurement as being accurately classified if the sensor measurement lies within the voxel boundaries of the voxel at which the maximum likelihood generating surface,i.e., the distance \( \arg \max \omega(s)\) is found.
On reaching the map boundary, if the visibility is not reduced sufficiently by the map volume, then the generating surface is determined to be accurate if the sensor measurement is greater than the distance to the map boundary, i.e., if \( \mathcal{Z}_r > s_\infty \).

%% file: implementation.tex
\subsection{Belief Propagation}
We initialise all the occupancy variables with a Bernoulli initial distribution as in~\cite{osman2016patches} with a probability of \(0.1\). The lower value corresponds to the observation that regions of interest are mostly empty. Lower prior probabilities aid in the initial inference steps since it permits the visibility terms to not saturate the outgoing messages. At inference time all keyframe images are ray traced and each ray atomically updates the corresponding nodes it traverses. We explicitly take into account the traversed length of each ray within a voxel to determine the contribution of the specific ray. The ray traversal is done via the standard 3DDA traversal algorithm~\cite{amanatides1987fast}. These are then accumulated to obtain the current incoming belief and prepared for the next pass of inference. Since storing all incoming messages for each voxel is prohibitive, we make an assumption that all rays going through a voxel from a single keyframe send the same average message. Thus, for each new keyframe image we create an auxiliary buffer that stores the average outgoing message. For the purpose of this work we perform three sequential up and down passes of belief propagation between all the factors and nodes respectively which we have empirically observed to be sufficient for obtaining convergence.

\subsection{Compact GPU representation}\label{sec:gvdb}
We extend and utilise the GVDB library~\cite{hoetzlein2016gvdb}, which represents a VDB~\cite{museth2013vdb}-like sparse data structure on GPUs. The smallest unit of allocated storage is a brick, which is a configurable size. In our implementation we chose it to be \(8^3\). Each level of the hierarchy then can contain a configurable number of bricks. Brick data is stored in contiguous space, enabling sparse data storage. The library allows for empty skip 3DDA ray traversals for regions where no data is stored. Fig.~\ref{fig:continuous} demonstrates a pictorial 2D example of a \(<2,2,2>\) configuration of the topology.

When adding new depth images, new bricks are allocated in a region around the reprojected 3D depth point cloud. This enables fast skip steps until within the neighbourhood of the ray measurement where the sensor model provides the most information. Once within an occupied brick, we traverse each voxel until a \(3\sigma \) distance beyond the measured depth at which point the sensor model provides no more useful information.

\subsection{Incorporating learnt sensor noise models}\label{sec:noise_depth}
Projective Infra-Red (IR) based RGB-D cameras and stereo cameras have a distinct quadratic increase in measurement uncertainty as a function of depth and often also have bias errors~\cite{basso2018robust}. We choose to model the sensor bias and noise as a function of distance travelled along a ray. Other factors such as angle of incidence, reflectivity, and texture are also some of the primary contributors to sensor noise that are only partially addressed within this model and are out of scope to model in this work. Within our implementation for the forward sensor model (Eq.~\ref{eq:noise}), at any given traversed length, we first utilise the sensor bias to predict what the mean sensor measurement would be and then evaluate the probability of a given measurement originating from this predicted measurement and distributed using the learnt noise value. Note that this is a significant difference to traditional approaches, where undistortion functions need to be computed~\cite{basso2018robust} as a first step that are then provided to conventional mapping processes. In addition to incorporating it within the mapping process to reduce latency, directly using the learnt sensor noise characteristics can also easily account for other sensor or environment idiosyncrasies without doing any additional preprocessing.

\begin{figure}[h]
  \centering
  \includegraphics[width=0.44\linewidth]{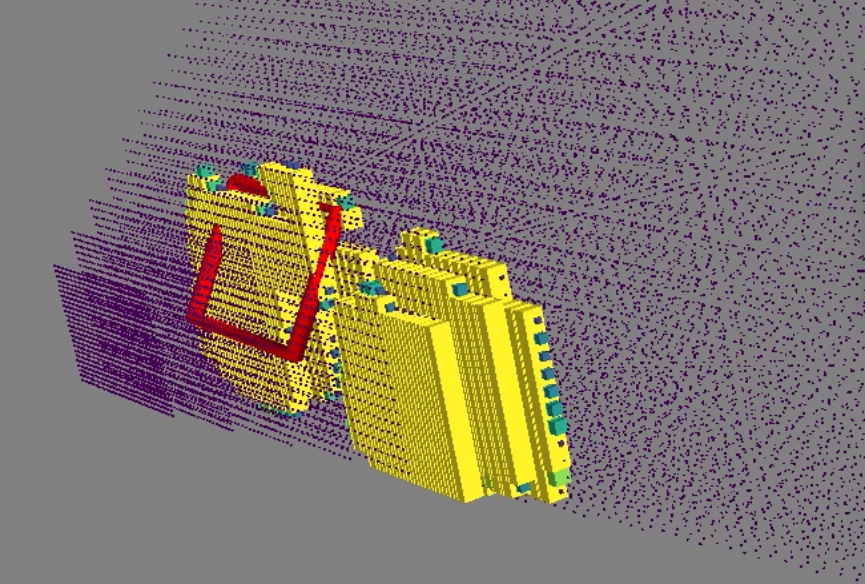}
  \includegraphics[width=0.5\linewidth]{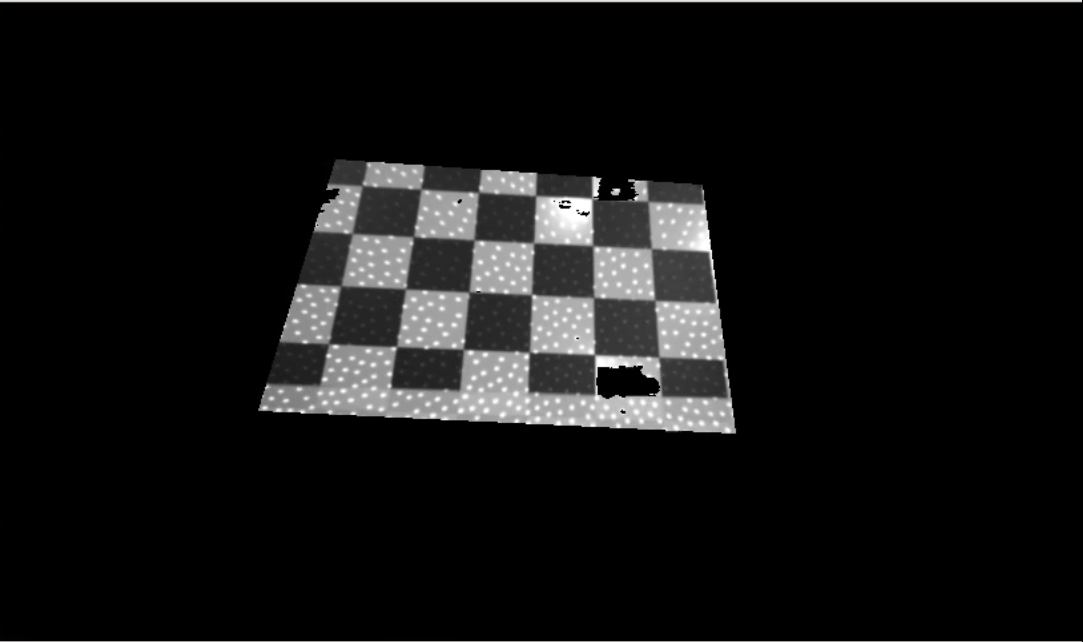}
  \includegraphics[width=0.9\linewidth]{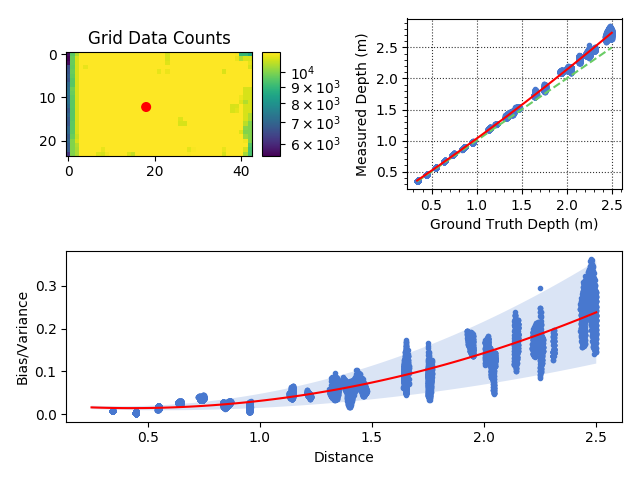}
  \caption{Top: A real-time utility that visualises the mocap calibration target location and the relevant slices of the view frustum for which sufficient data has been collected for learning the forward sensor noise model. Bottom: After collecting up to the order of \(10^4\) data points for each valid \(20 \times 20\) pixel cell, the measured depth is regressed against the ground truth depth to fit a polynomial sensor noise model and the residuals are used to regress the bias and variance. Shown for a Realsense D435 operating at \(848\times480\)~px resolution.}\label{fig:utility}
\end{figure}
\begin{figure}[h]
  \centering
  \includegraphics[width=0.49\linewidth]{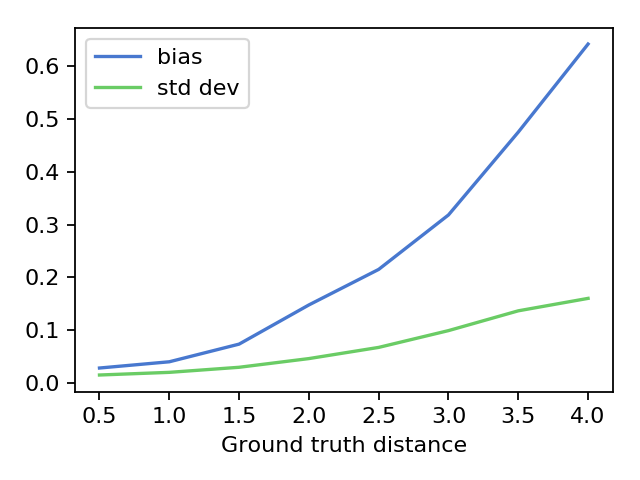}~\includegraphics[width=0.49\linewidth]{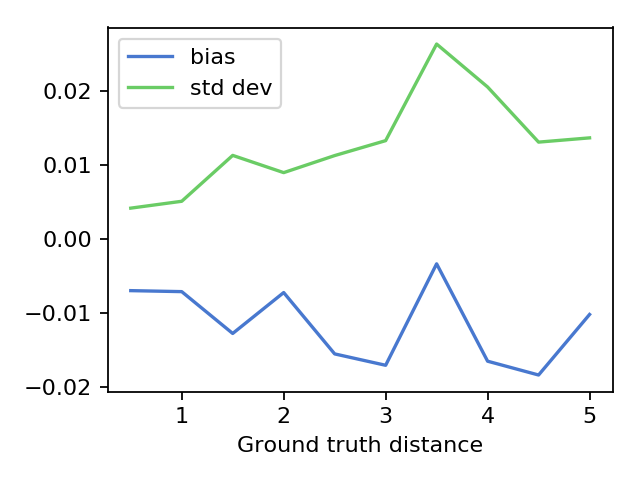}
  \caption{Meta level characteristics of the bias and standard deviation for the Realsense D435 and the Kinect One on the left and right, respectively. Note that the characteristics of the D435 camera follow a quadratic profile whereas those of the Kinect One are almost linear.
  }\label{fig:meta}
\end{figure}

We thus choose to directly learn the sensor bias and noise for every pixel region for all depths in the operating range of the camera. We collect measured and ground truth depths for pixel regions and fit a \(2^{\text{nd}}\) degree polynomial as suggested by prior literature~\cite{basso2018robust}. These coefficients are then loaded into GPU memory and are used during inference and accuracy estimation (see Fig.~\ref{fig:utility}).

\subsubsection{Instrumentation}
Motion capture (mocap) is used to determine camera and a calibration target's extrinsics. The capture volume adequately covers the sensor frustum for both the camera sensors. Next, temporal lag between the motion capture data stream and the image stream is computed by comparing the relative velocities of the calibration target within the image frame and within the mocap arena.
\subsubsection{Acquisition Process}
The camera is held static on a tripod while a chessboard is moved around in the view frustum of the depth camera. The projected calibration target dimensions in the image are used to create a valid masked depth image. For every valid pixel in the masked depth image, the corresponding ground truth location is determined by performing a ray to plane intersection test using the relative transform obtained from the relevant mocap frames. Thus, each masked depth image provides a large number of ground truth and measured depth values per pixel location. This data is then binned over small configurable pixel neighbourhoods that are then used to fit bias and noise models (Fig.~\ref{fig:utility}). We choose a \(20 \times 20\) pixel neighbourhood to trade-off noise model fidelity and memory usage on the GPU. Overall meta trends of the bias and variance of the raw data compared to the ground truth are shown in Fig.~\ref{fig:meta}. These are generated by taking the mean of the bias and standard deviation images in the central third patch of pixels and are used in our evaluation.

%% file: results.tex
The emphasis of our results here is that
\begin{itemize}
    \item The MRFMap framework enables higher fidelity mapping than OctoMap~\cite{hornung2013octomap}, the standard robotic grid based volumetric occupancy framework, especially for noisy data; and
    \item The time taken for ray tracing and inference is much faster than CPU based approaches, especially at finer map resolutions.
\end{itemize}

Quantitatively, we measure the accuracy of the maps by utilising the metric discussed in Sec.~\ref{sec:metric}. All pixels (rays) for which the distance of the maximum likelihood generating surface, \(s\) lies within a constant \(1.5 \sigma \) standard deviation of the measured depth are classified as accurate, and those that are not are classified as being incorrect. Sample accuracy images are shown in Fig.~\ref{fig:noise_sim}. The final accuracy score for a given depth image is then the sum of all pixels (rays) that are accurately explained by the map over the total number of valid depth pixels in the depth image. The accuracy score is then computed over all scans in the dataset and the mean and the standard deviation reported in the tables below.

We detail three different scenarios
\begin{enumerate}
    \item A simple simulated scene in the Gazebo\footnote{\url{http://gazebosim.org/}} simulator to demonstrate the accuracy measure and ability to encode bias and noise modelling within inference;
    \item A standard open access dataset with very precise simulated sensor noise for RGB-D sensors to demonstrate inference with simulated real-world noise; and
    \item A real-world dataset where we use a noisy sensor (Realsense D435) and evaluate the generated map using a much more accurate sensor (Kinect One).
\end{enumerate}

\begin{figure}[h]
    \centering
    \includegraphics[width=0.45\linewidth]{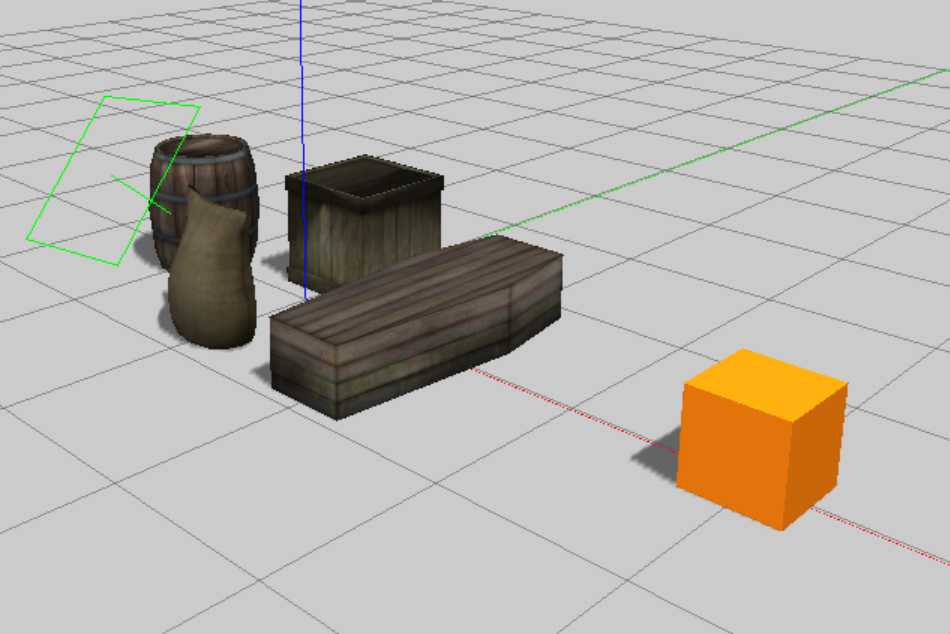}~\includegraphics[width=0.45\linewidth,clip,trim={400 200 700 900}]{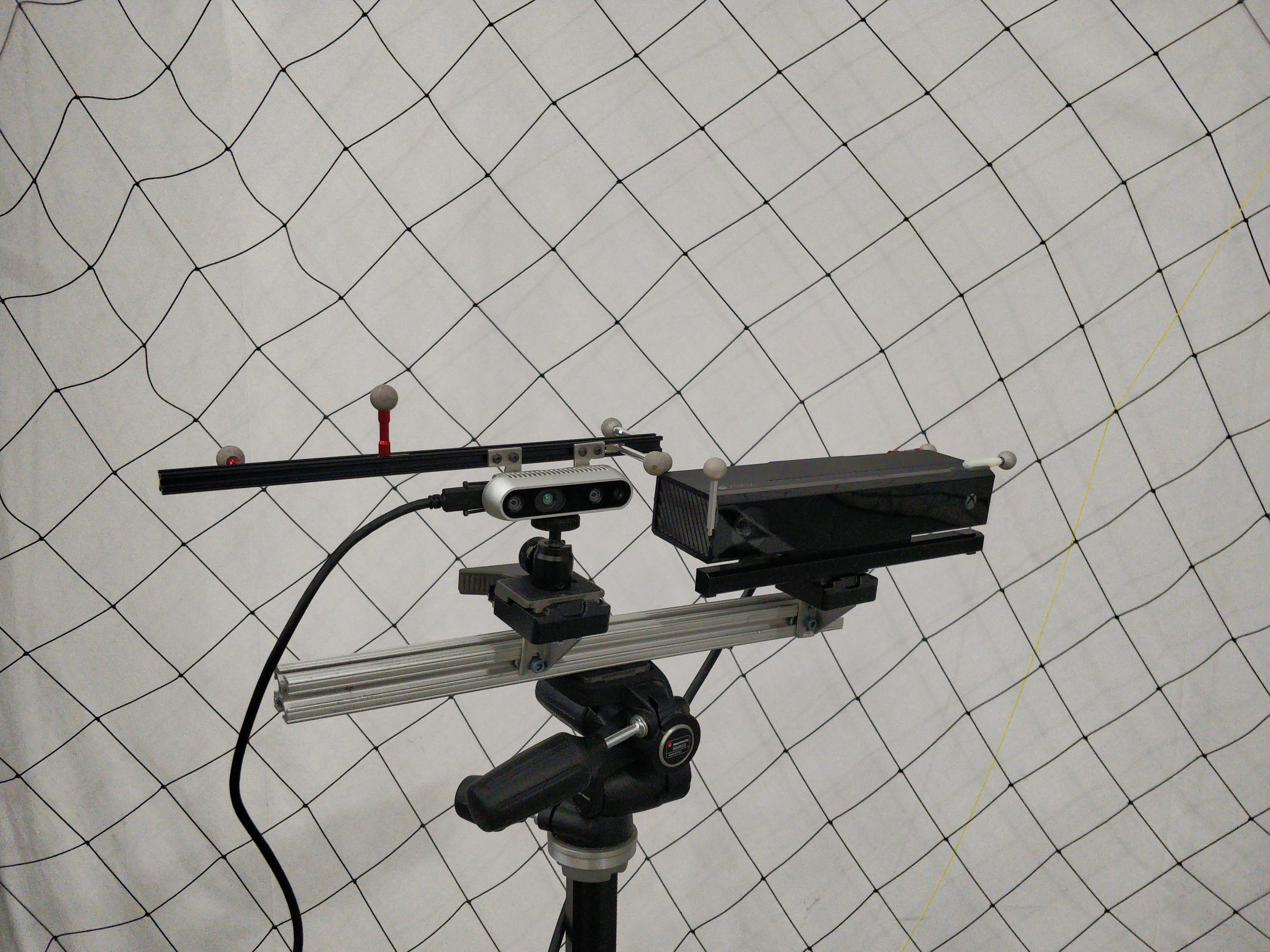}
    \caption{Left: We use a synthetic scene in the Gazebo simulator as viewed from 12 different simulated depth camera views of \(640 \times 480\) px resolution. Right: Camera rig for a Realsense D435 and a Kinect One sensor used for acquiring the real-world dataset.}\label{fig:sim_reconstruct}
\end{figure}

\subsubsection{Demonstrating the accuracy metric}
\begin{figure}[h]
    \centering
        \includegraphics[width=\linewidth]{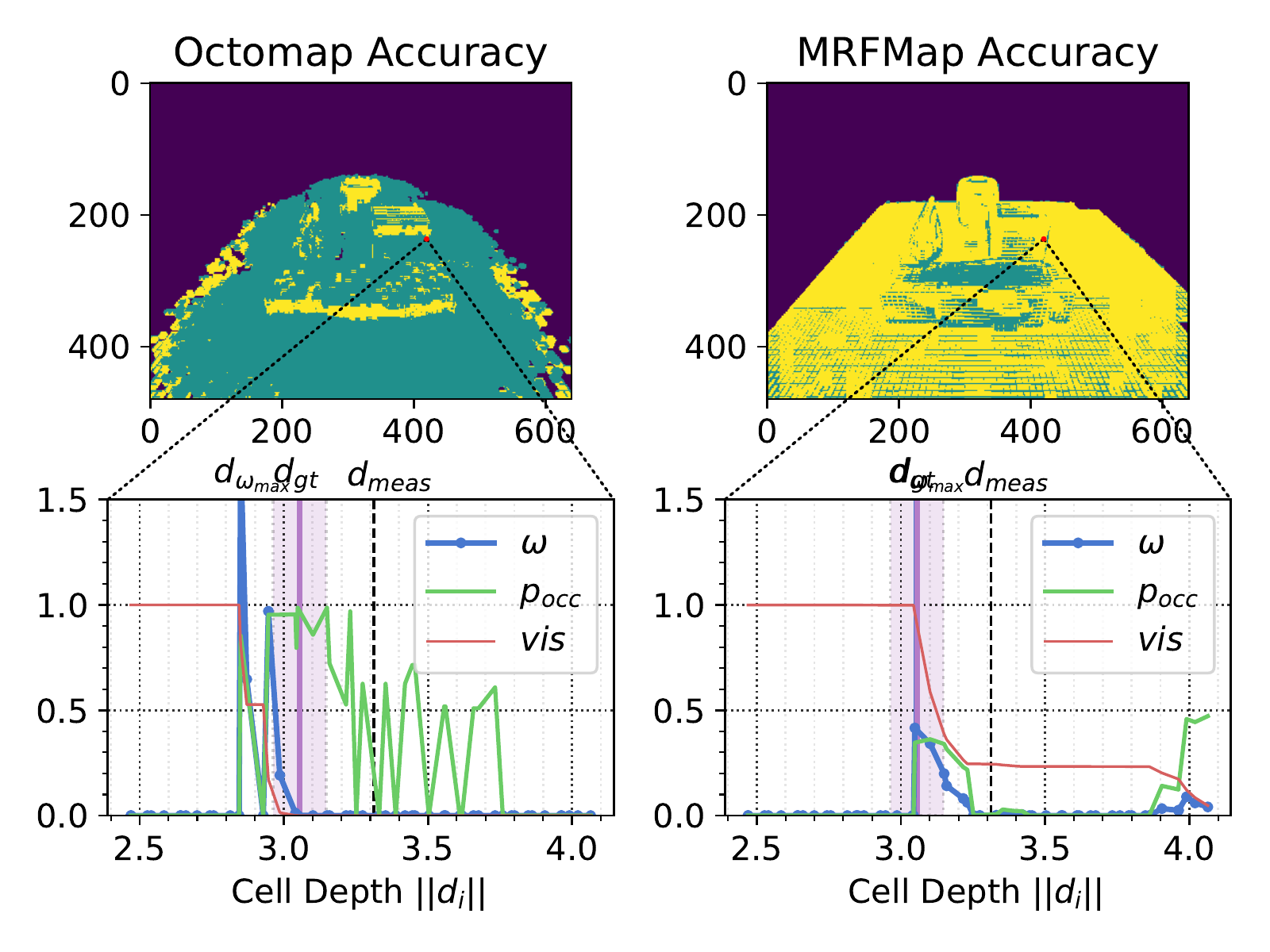}
        \caption{Computing accuracy with simulated learnt noise. Top: Accuracy images. Correctly predicted pixels are coloured yellow, incorrectly predicted pixels as green, and invalid pixels as purple. Insets: Plots of ray-traced \(p_{occ}(s), \omega(s), \text{and} \text{vis}(s)\) for the selected pixel. Purple vertical line represents the ground truth depth \(d_{gt}\), with a corresponding \(1 \sigma \) band around it to determine if the maximum likelihood generating surface location is present within. Dashed vertical line is the depth measurement \(d_{meas}\) at the selected pixel in this held out sampled depth image for evaluation. Blue vertical line is the location of the maximum likelihood generating surface \(d_{max} = \arg \max \omega(s) \)(Sec.~\ref{sec:metric}). Note how for the selected ray at the right edge of the cuboidal box, the MRFMap accurately describes the probability of occupancy along the ray (green). The ray first hits the cuboidal box, corresponding to the first peak of \(\omega\) around \(3.05~\text{m}\) and then travels through empty space till it hits the ground at a distance of \(4.0~\text{m}\). For the OctoMap, an incorrect over-confident map occupancy probability causes the visibility (red) to drastically drop and it incorrectly predicts the most likely generating surface.}\label{fig:noise_sim}
    
\end{figure}

For the Gazebo simulation environment shown in Fig.~\ref{fig:sim_reconstruct} we obtain 12 ground truth ray traced images and then sample simulated depth images by utilising the learnt per \(20 \times 20~\text{px}\) patch bias and standard deviation noise models for the Realsense D435 at \(640 \times 480~\text{px}\) resolution. These simulated true noise images are used to infer the probabilistic maps and are then evaluated for accuracy in a leave one out cross-validation scheme. A sample view is shown in Fig.~\ref{fig:noise_sim} and the corresponding accuracy is shown in Table~\ref{tab:acc_sim}. For a specific pixel, we plot the occupancy probability of the map \(p_{occ}\), the visibility \(\text{vis}\), and the generating surface probability \(\omega(s) \). The peak of the \( \omega(s) \) distribution is highlighted by a blue vertical line - thus, according to the accuracy metric the pixel would be classified as being accurate if the ground truth depth measurement at that pixel lies within \(1\sigma \) of this distance. This region is shown as a purple translucent region. As can be seen in Fig.~\ref{fig:sim_reconstruct} for the given pixel the MRFMap predicted ground truth distance lies within this region, while that for the OctoMap does not. The dashed vertical line is the noisy measurement sampled from the ground truth for the pixel that was used for inferring the map - highlighting the fact that incorporating the forward sensor noise and the bias within the inference allows to compensate for noisy, inflated depth measurements.

Since the injected depth noise and bias is non-trivial, OctoMap struggles at accurately inferring the map, while MRFMap infers the map accurately, predicting the generating surface location for each pixel to be close to the ground truth depth, thus demonstrating the ability to compensate for injected bias and noise when correctly modelled.

\begin{table}[h]
    \begin{center}
        \begin{tabular}{lccc}
            \toprule
             Resolution   &         0.01m         &         0.02m         &         0.05m         \\
            \midrule
             MRFMap        & \( 0.870 \pm 0.011 \) & \( 0.917 \pm 0.022 \) & \( 0.843 \pm 0.017 \) \\
             OctoMap       & \( 0.082 \pm 0.010 \) & \( 0.118 \pm 0.020 \) & \( 0.213 \pm 0.049 \) \\
            \bottomrule
            \end{tabular}
    \end{center}
    \caption{Accuracy values for the simulated ground truth environment using leave one out cross-validation over 12 images in a \(4\times4\times3~\text{m}^3\) environment.}\label{tab:acc_sim}
\end{table}

\subsubsection{Demonstration on publicly available dataset with known ground truth}
We use the \texttt{livingroom1} noisy depth sequence of the Augmented ICL-NUIM Dataset~\cite{Choi_2015_CVPR} since it accurately models sensor noise characteristics of projective IR based RGB-D cameras. To select the keyframes to perform inference we use a simple geometric displacement based heuristic. A new keyframe is generated if the tangent norm of the translation or the rotation from the last keyframe pose is larger than a threshold (here 1.0 for each). To evaluate the accuracy we then use the ground truth depth images from the same trajectory. We use the Kinect noise model as reported in ~\cite{basso2018robust} for performing map inference, and a constant noise model for the evaluation. Accuracy results are presented in Table~\ref{tab:acc_icl}. 
As expected, at finer resolutions, the noise model can very precisely impact inference involving multiple voxels and thus provides much more accuracy than at lower voxel resolutions.

\begin{figure}[h]
    \begin{center}
        \includegraphics[width=0.45\linewidth]{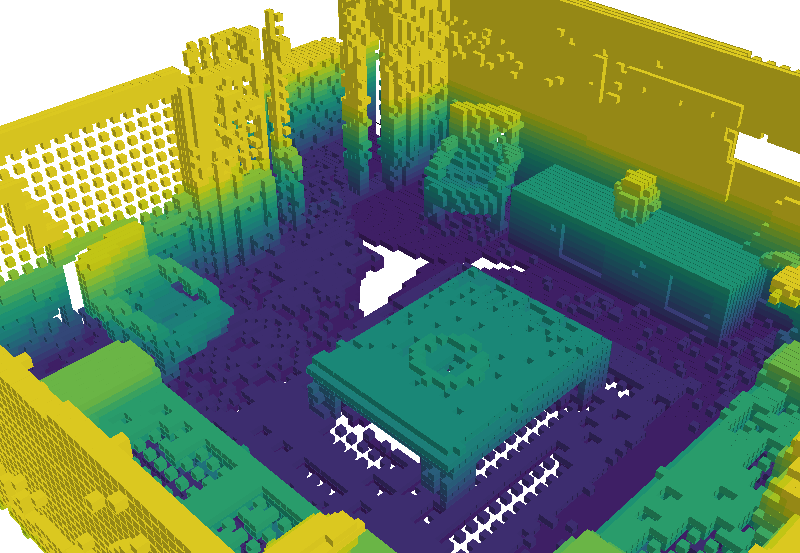}
        \includegraphics[width=0.45\linewidth]{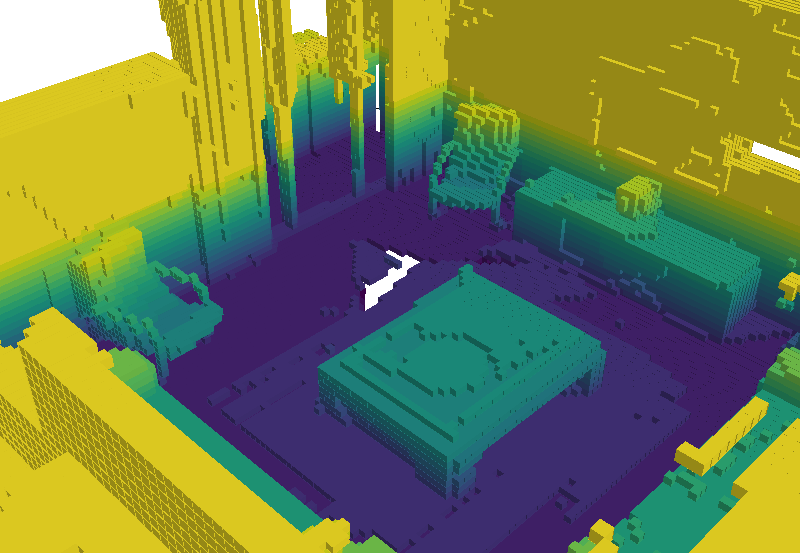}
    \end{center}
    \caption{Occupied voxel map for the \texttt{livingroom1} noisy depth sequence of the Augmented ICL-NUIM dataset~\cite{Choi_2015_CVPR}. \(10\times10\times2~\text{m}^3\) volume at \(0.05~\text{m}\) resolution. Left: OctoMap, Right: MRFMap. Note the artifacts in the OctoMap due to a combination of glancing rays and sensor noise.}\label{fig:ICL}
\end{figure}
\begin{table}[h]
    \begin{center}
        \begin{tabular}{lccc}
            \toprule
            Resolution & \(0.01~\text{m}\)     & \(0.02~\text{m}\)     & \(0.05~\text{m}\)     \\
            \midrule
            MRFMap      & \( 0.945 \pm 0.097 \) & \( 0.939 \pm 0.101 \) & \( 0.891 \pm 0.114 \) \\
            OctoMap     & \( 0.922 \pm 0.105 \) & \( 0.882 \pm 0.108 \) & \( 0.723 \pm 0.113 \) \\
            \bottomrule
        \end{tabular}
    \end{center}
    \caption{Accuracy means and standard deviations over the entire dataset for the \texttt{livingroom1} noisy depth sequence. The MRFMaps are constructed using the Kinect forward sensor noise model and evaluated on a constant noise model. Map volume is \(10\times10\times5~\text{m}^{3}.\)}\label{tab:acc_icl}
\end{table}

\subsubsection{Real-world dataset}
\begin{figure}[h]
    \centering
    \begin{subfigure}[b]{\columnwidth}
        \centering
        \includegraphics[width=0.4\linewidth]{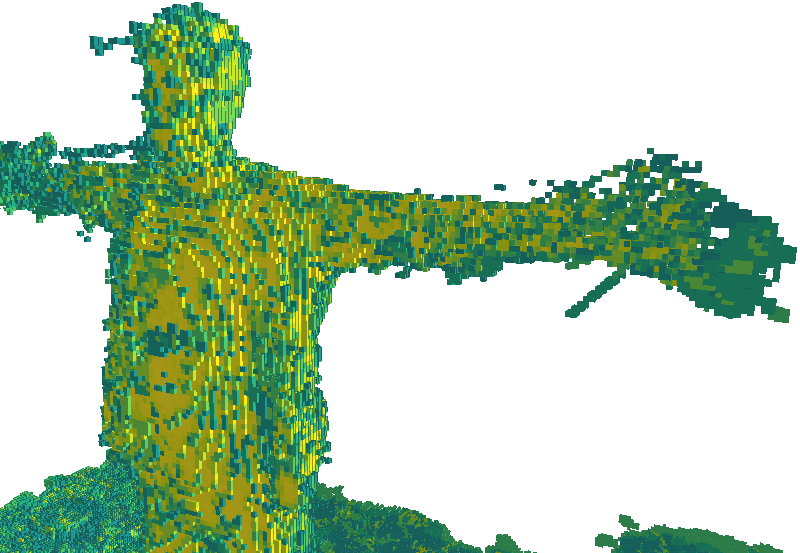}
        \includegraphics[width=0.4\linewidth]{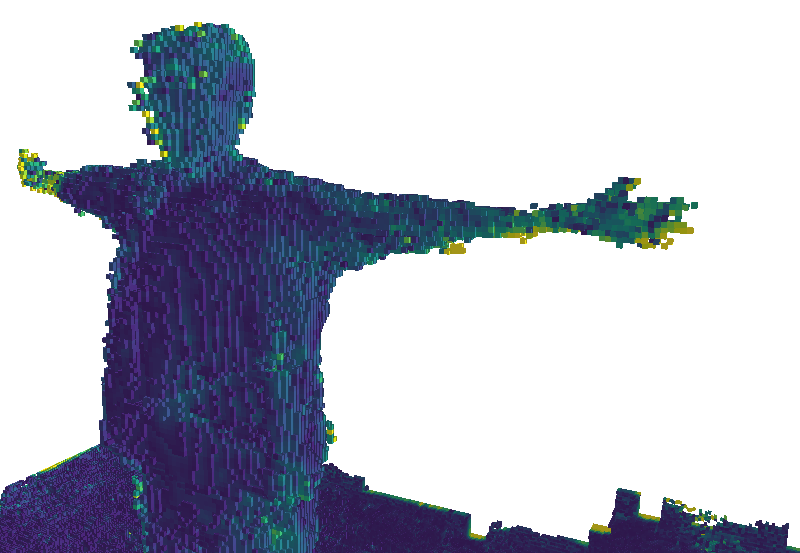}
        \caption{\(0.01~\text{m}\)}
    \end{subfigure}
    \hfill
    \begin{subfigure}[b]{\columnwidth}
        \centering
        \includegraphics[width=0.4\linewidth]{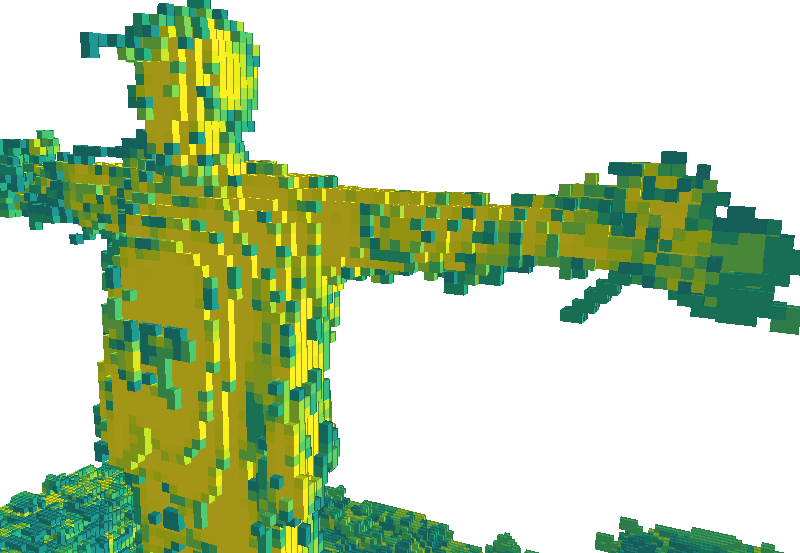}
        \includegraphics[width=0.4\linewidth]{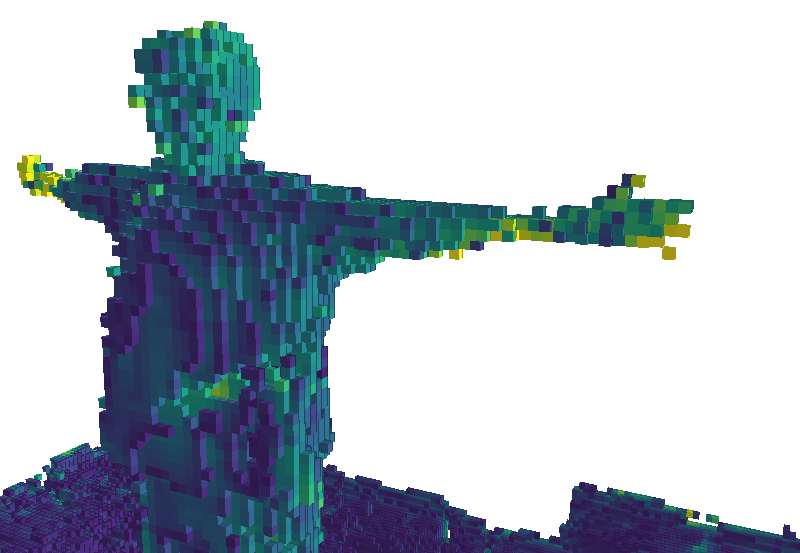}
        \caption{\(0.02~\text{m}\)}
    \end{subfigure}
    \hfill
    \begin{subfigure}[b]{\columnwidth}
        \centering
        \includegraphics[width=0.4\linewidth]{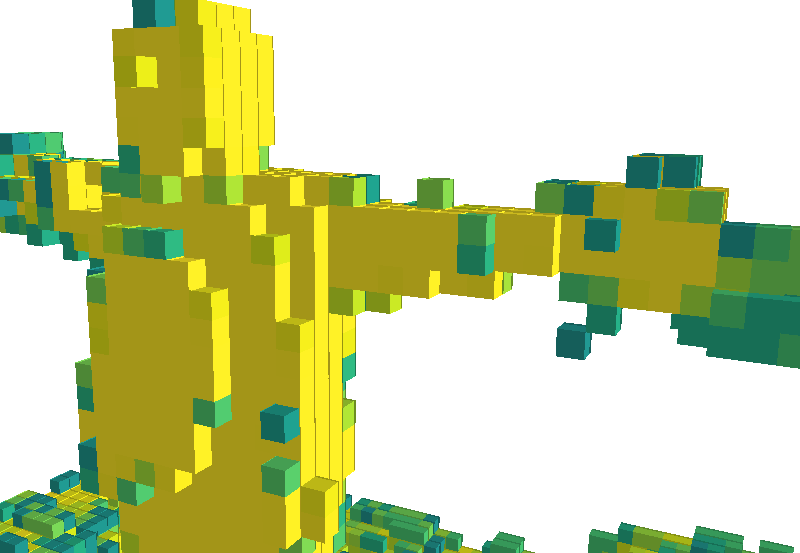}
        \includegraphics[width=0.4\linewidth]{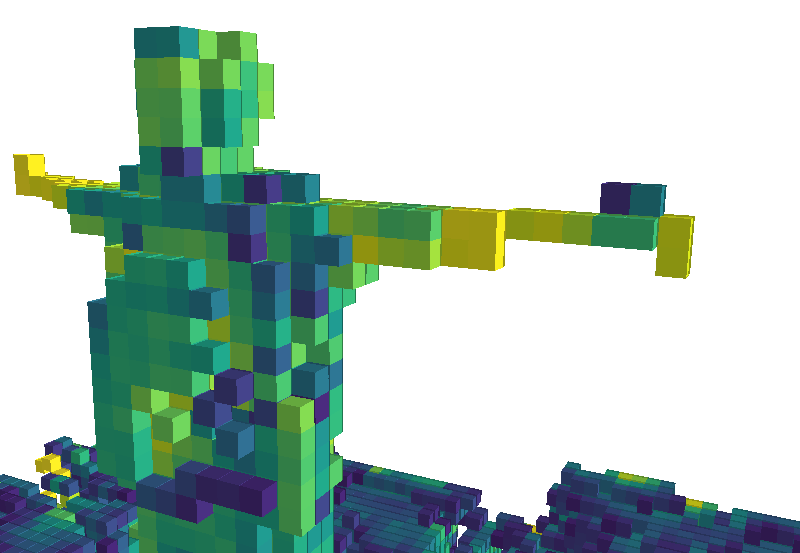}
        \caption{\(0.05~\text{m}\)}
    \end{subfigure}

    \caption{OctoMap (left) and MRFMap (right) at various map resolutions on real-world data. The MRFMaps are constructed using Realsense D435 data at \(848\times480~\text{px}\) resolution with the aggregate forward sensor noise model (including bias) shown in Fig.~\ref{fig:meta} and evaluated on a constant noise model using a rigidly attached Kinect One. Note the much better reconstructed fidelity of the MRFMaps in the head and the hand region. Dark violet represents lower occupancy probability while bright yellow represents high occupancy probability.}\label{fig:real_world_aditya}
\end{figure}

For the real-world dataset we utilise a rig shown in Fig.~\ref{fig:sim_reconstruct} in a Vicon motion capture arena and capture a human subject. Maps are built with the noisier stereo IR Realsense D435 camera and accuracy evaluated with data from the Kinect One, a Time of Flight (ToF) based camera. The latter is employed for evaluating accuracy in the absence of having ground truth depth. For inference, we use the aggregate polynomial model for bias and standard deviation demonstrated in Fig.~\ref{fig:meta} for all pixels. The keyframes are selected based on the same geometric heuristic with smaller thresholds (0.5 each) to account for the smaller capture volume. Low accuracy values can be attributed to errors in the inter-camera registration and the dynamic camera motion. Inferred maps are shown in Fig.~\ref{fig:real_world_aditya} at multiple resolutions and highlight the drastically better qualitative maps obtained from the noisy sensor data.

\begin{table}[h]
    \begin{center}
        \begin{tabular}{lccc}
            \toprule
            Resolution & \(0.01~\text{m}\)     & \(0.02~\text{m}\)     & \(0.05~\text{m}\)     \\
            \midrule
            MRFMap      & \( 0.305 \pm 0.185 \) & \( 0.335 \pm 0.195 \) & \( 0.406 \pm 0.188 \) \\
            OctoMap     & \( 0.309 \pm 0.192 \) & \( 0.302 \pm 0.191 \) & \( 0.263 \pm 0.170 \) \\
            \bottomrule
        \end{tabular}
    \end{center}
    \caption{Accuracy values for the real-world dataset shown in Fig.~\ref{fig:real_world_aditya}. Map volume is \(3\times3\times2~\text{m}^{3}.\)}\label{tab:acc_aditya}
\end{table}

\subsubsection{Timing}
\begin{figure}[h!]
    \begin{center}
        \includegraphics[width=0.9\linewidth]{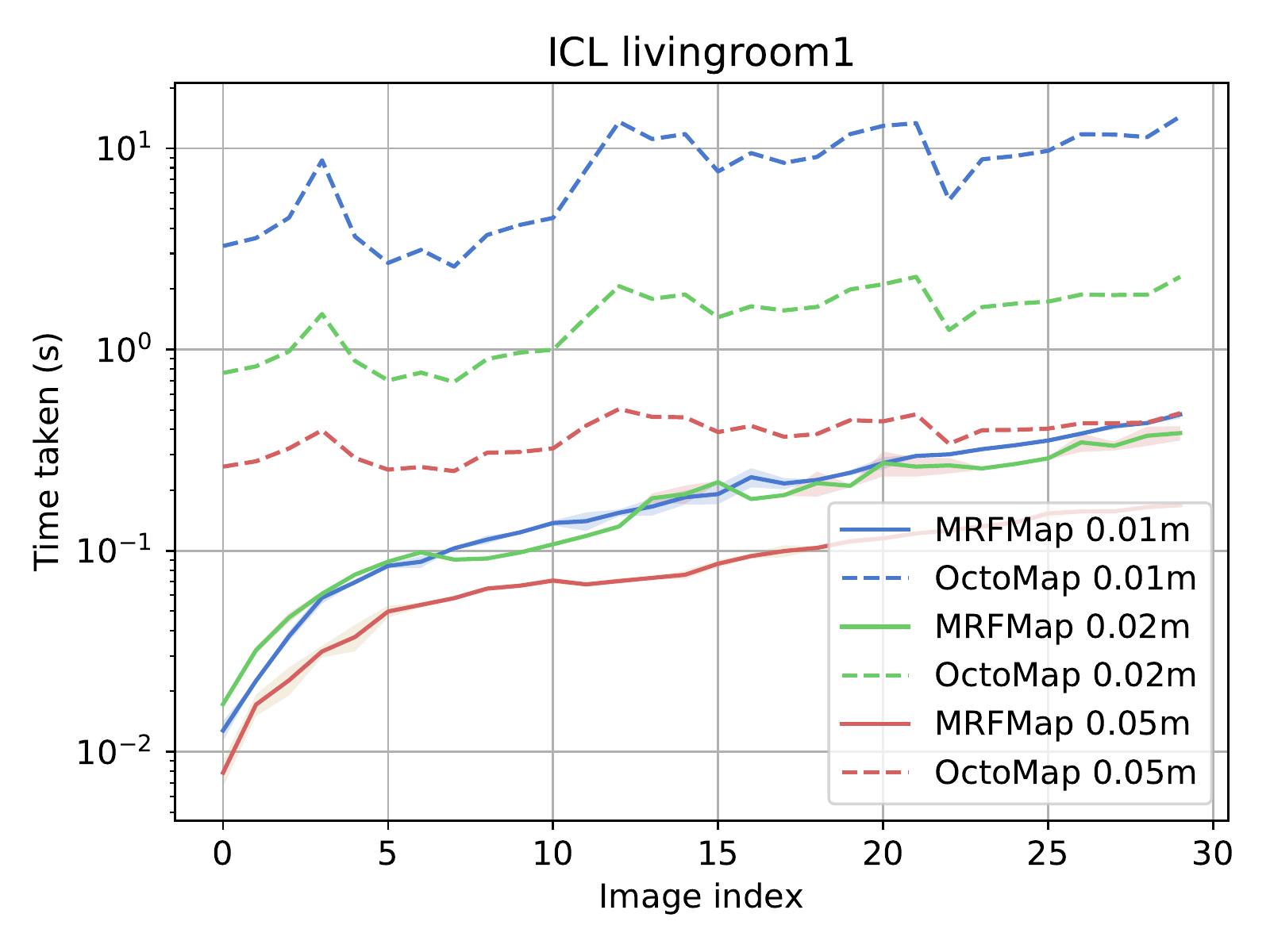}
        \caption{Incremental time taken for adding a new image and performing inference in an MRFMap and an OctoMap to generate Fig.~\ref{fig:ICL}}\label{fig:timing_ICL}
    \end{center}
\end{figure}

By virtue of being a framework that retains keyframe sensor data in memory,  adding a new image causes the inference to iterate over all the images, and the overall time taken for inference increases monotonically. However, due to the accelerated data structure and parallelised implementation, an MRFMap is much faster at ray tracing and performing map inference than an OctoMap, especially at finer resolutions. Fig.~\ref{fig:timing_ICL} demonstrates that, for instance, at a resolution of \(0.01~\text{m}\) on the \texttt{livingroom1} dataset even after adding 30 keyframes, ray tracing a new image and performing three passes of inference over all keyframes still takes two orders of magnitude less time than adding it to the OctoMap. These results were obtained on an NVIDIA RTX 2060 Super and an eight-core AMD Ryzen 3700x CPU.

%% file: conclusions.tex
Through our experiments, we show that explicitly modelling the intra-and inter-dependence of neighbouring voxels due to sensor ray formation as opposed to treating all voxels as being independent enable more accurate occupancy inference than conventional occupancy grid mapping frameworks. Incorporating learnt forward sensor models helps obtain even higher map fidelity, and ensures that the map is a maximum likelihood solution that best attempts to explain the sensor data.

The MRFMap framework is envisioned as an essential building block for replacing traditional grid based submapping implementations~\cite{estrada2005hierarchical,ni2007tectonic,sodhi2019online} since it enables performing on-demand inference to obtain a probabilistically accurate map without having to approximately resample the submaps when moving any subset of keyframe poses. The monotonically increasing memory and runtime considerations can be addressed by marginalising out occupancy in the inactive regions. In the future we intend to explore these two capabilities for performing dense probabilistic SLAM.

%% file: supplement.tex
\pagebreak
\onecolumn

\begin{center}
\textbf{Supplementary Material to:\\ MRFMap: Online Probabilistic 3D Mapping using Forward Ray Sensor Models}\\
{Kumar Shaurya Shankar and Nathan Michael}\\
\textit{This document derives the messages for the sum-product belief propagation algorithm presented in the paper and should be treated as an appendix to the same.}
\end{center}

\setcounter{equation}{0}
\setcounter{figure}{0}
\setcounter{table}{0}
\setcounter{section}{0}
\setcounter{page}{1}
\renewcommand{\theequation}{S\arabic{equation}}
\renewcommand{\thefigure}{S\arabic{figure}}
\renewcommand{\bibnumfmt}[1]{[S#1]}
\renewcommand{\citenumfont}[1]{S#1}

\section{Sum-Product Belief Propagation}

\subsection{Sum-Product Belief Propagation}
Sum-Product belief propagation is a common message-passing algorithm for performing inference on factor graphs. By exploiting marginalisation of joint distributions using factorisation of a graph it enables computing marginal distributions very efficiently. A factor graph is a bipartite graph containing nodes corresponding to variables and factors that are connected by edges. Messages are passed between connected nodes and factors that try to influence the marginal belief of their neighbours. The passing continues until convergence (if any) is achieved.

The message sent from a variable node \(x\) to a factor \(f\) is the cumulative belief of all the incoming messages from factors to the node except the factor in question
\begin{equation}
    \mu_{x \rightarrow f}(x) = \prod_{g \in \mathcal{F}_x  \backslash f} \mu_{g \rightarrow x}(x) \label{eq:node_to_factor},
\end{equation}
where \(\mathcal{F}_x\) is the set of neighbouring factors to \(x\). Similarly, the message sent from the factor to the node is the marginalisation of the product of the value of the factor \(\phi_f\) with all the incoming messages from nodes other than the node in question
\begin{align}
    \mu_{f \rightarrow x}(x) = \sum_{\mathcal{X}_f \backslash x} \phi_f(\mathcal{X}_f) \prod_{y \in \mathcal{X}_f \backslash x} \mu_{y \rightarrow f}(y), \label{eq:factor_to_node}
\end{align}
where \(\mathcal{X}_f\) is the set of all neighbouring nodes of \(f\).

Upon convergence, the estimated marginal distribution of each node is proportional to the product of all messages from adjoining factors
\begin{equation}
    p(x) \propto \prod_{g \in \mathcal{F}_x} \mu_{g \rightarrow x}(x)
\end{equation}
Similarly, the joint marginal distribution of the set of nodes belonging to one factor is proportional to the product of the factor and the messages from the nodes
\begin{equation}
    p(\mathcal{X}_f) \propto \phi_f(\mathcal{X}_f) \prod_{\mathcal{X}_f} \mu_{x \rightarrow f}(x)
\end{equation}

\subsection{Markov Random Field}
Each ray in all the cameras generates a factor graph, the joint distribution of which is
\begin{equation}
    p( \Oc, \mathbf{\D} ) = \frac{1}{Z} \prod_{i \in \mathcal{X}} \phi_i(o_i) \prod_{r \in \mathcal{R}} \psi_r(\Oc_r, \D_r),
\end{equation}
where \(\mathcal{X}\) is the set of all the voxels \(o_i \in \{0,1\}\) and \(\mathcal{R}\) is the set of all the rays from all the cameras viewing the scene, \(\Oc_r = \{o_1^r, \dots, o_{N_r}^r\}\) is the list of all the voxels traversed by a ray \(r\), \(\D_r\) is its corresponding depth variable, and \(Z\) is the normalisation constant. The total set of all the occupancy and depth variables are summarised as \(\Oc = \{o_i \mid i \in \mathcal{X}\}\) and \(\mathbf{d} = \{d_r \mid r \in \mathcal{R}\}\). \(\phi_i\) and \(\psi_r\) are the potential factors as described as follows:

\subsubsection{Prior Occupancy Factor}
This is simply a unary factor assigning an independent Bernoulli prior \(\gamma\)to the voxel occupancy label for each voxel
\begin{equation}
    \phi_i(o_i) = \gamma^{o_i}(1 - \gamma)^{1 - o_i}.
\end{equation}
Note that these priors can be informed from predictive methods if required.

\subsubsection{Ray Depth Potential Factor}
A ray potential creates a factor graph connecting the binary occupancy label \(o_i\) for all voxels traversed by a single ray by virtue of making a measurement. Each of these voxels has a corresponding distance from the camera origin, and thus the ray is defined to include a depth variable \(\D_r\) that represents the event that the measured depth is at distance \(d_{i}^r\). For this to happen all the preceding voxels ought to be empty and the corresponding voxel needs to be occupied. This leads to the definition of the joint occupancy and depth potential factor
\begin{equation}
    \psi_r(\Oc_r,\D_r ) = \begin{cases}
        \nu_r(d_{i}^r) & \text{if } d_r = \sum_{i=1}^{N_r} o_i^r \prod_{j<i} (1 - o_j^r)  d_i^r \\
        0              & \text{otherwise}
    \end{cases}.\label{eq:depth_potential}
\end{equation}
Here \(\nu_r(d_{i}^r)\) denotes the probability of observing depth \(d_{i}^r\) given the measured depth value \(\mathcal{D}^r\) that we model as \(\nu_r(d_{i}^r) = \mathcal{N}(d_{i}^r; \mathcal{D}_r, \sigma(d_{i}^r)) \). Note that we model this probability to vary in mean and noise as a function of the distance from the camera, which enables utilising learnt sensor noise characteristics.

This ray potential measures how well the occupancy and the depth variables explain the depth measurement \(\mathcal{D}_r\). This is more apparent when Eq.~\ref{eq:depth_potential} is written as
\begin{equation} \label{eq:depth_potential_cases}
    \psi_r(\Oc_r,\D_r ) = \begin{cases}
        \nu_r(d_1^r) & \text{if}~ d^r = d_1^r, o_1^r = 1                                       \\
        \nu_r(d_2^r) & \text{if}~ d^r = d_2^r, o_1^r = 0, o_2 = 1                              \\
        \vdots       &                                                                         \\
        \nu_r(d_N^r) & \text{if}~ d^r = d_N^r, o_1^r = 0, \dots, o_{{N_r}-1}^r = 0, o_{N_r}^r  = 1
    \end{cases}
\end{equation}
This sparse structure of the ray potential enables massive simplification of the message passing equations, as shown next.

\section{Message Passing Derivation}
\subsection{Ray Depth Potential to Depth Variable Messages}
Since we're only concerned about the depth variable, we marginalise out the messages from all the occupancy nodes going to the ray depth potential. Following~\ref{eq:factor_to_node} we have
\begin{equation}
	\mu_{\psi_r \rightarrow \D_r}\left(\D_r=d_{i}^r\right)=\sum_{o_{1}^r} \dots \sum_{ o_{N_r}^r} \psi_r\left(\mathbf{o}_r, \D_r\right) \prod_{j=1}^{N_r} \mu_{o_j^r \to \psi_r}\left(o_{j}^r\right).
\end{equation}
Na\"ively evaluating this equation is not feasible. However, we can exploit the sparse diagonal nature of the ray depth potential to recursively simplify this expression.
After dropping the ray index for notational convenience and abbreviating \( \mu_{o_j^r \to \psi_r}(o_{j}^r)\) as \(\mu(o_j)\) we have
\begin{align}
    \mu_{\psi \rightarrow \D}\left(\D=d_i\right)
     & = \mu\left(o_{1}=1\right)  \overbrace{\left[\sum_{o_{2}} \dots \sum_{o_{N}} \psi\left(o_{1}=1, o_{2}, \dots, o_{N}, d=d_{i}\right) \prod_{j=2}^{N} \mu\left(o_{j}\right)\right]}^{\bigtriangleup }  \nonumber \\
     & + \mu\left(o_{1}=0\right)\underbrace{\left[ \sum_{o_{2}} \dots \sum_{o_{N}} \psi\left(o_{1}=0, o_{2}, \ldots, o_{N}, d=d_{i}\right) \prod_{j=2}^{N} \mu\left(o_{j}\right) \right]}_{\Box }
\end{align}
From Eq.~\ref{eq:depth_potential_cases}, for the top expression \(\bigtriangleup\) the ray potential term \(\psi\left(o_{1}=1, o_{2}, \dots, o_{N}, d=d_{i}\right)\) evaluates to \( \nu(d_1) \) if \(i = 1\) and \(0\) otherwise. Since it only depends on \(d_1\) it can be brought out of the summation as follows:
\begin{equation}
    \bigtriangleup = \nu(d_1) \underbrace{\sum_{o_{2}} \dots \sum_{o_{N}}\prod_{j=2}^{N} \mu\left(o_{j}\right)}_{\text{evaluates to 1.}} .
\end{equation}
Assuming that all the incoming messages \(\mu\) are normalised such that they sum to 1, the terms highlighted with the underbrace evaluate to 1. We maintain this normalisation in our implementation.

Assuming that \(i \neq 1\), the bottom expression \(\Box\) can be recursively expanded similar to this step. Each such expansion brings in a term of the form \( \prod_{k<j} \mu(o_k=0) \mu(o_j=1) \nu{(d_j)}  \), until we reach the \(i\)th term Thus
\begin{align}
    \mu_{\psi \rightarrow \D}\left(\D=d_i\right)
     & =   \prod_{k<i} \mu(o_k=0) \bigg[ \mu(o_i=1)  \sum_{o_{i+1}}  \dots \sum_{o_{N}} \overbrace{\psi\left(o_{1}=0, o_{2}=0, \ldots, o_{i}=1, o_{i+1}, \dots o_{N}, d=d_{i}\right)}^{\text{evaluates to } \nu(d_i)}  \prod_{j={i+1}}^{N} \mu\left(o_{j}\right)  \nonumber \\
     & +\mu(o_i=0)  \sum_{o_{i+1}}  \dots \sum_{o_{N}}  \underbrace{\psi\left(o_{1}=0, o_{2}=0, \ldots, o_{i}=0, o_{i+1}, \dots o_{N}, d=d_{i}\right)}_{\text{evaluates to } 0}  \prod_{j={i+1}}^{N} \mu\left(o_{j}\right) \bigg]
\end{align}
where the first term evaluates to \(\nu(d_i)\), and the next term evaluates to 0, giving us
\begin{align}
    \mu_{\psi \rightarrow \D}\left(\D=d_i\right)
     & = \prod_{k<i} \mu(o_k=0) \mu(o_i=1) \nu{(d_i)}  \underbrace{\sum_{o_{i+1}}  \dots \sum_{o_{N}} \prod_{j={i+1}}^{N} \mu\left(o_{j}\right)}_{\text{evaluates to } 1} \nonumber \\
     & = \nu{(d_i)} \mu(o_i=1) \prod_{k<i} \mu(o_k=0)
\end{align}

\subsection{Depth Variable to Ray Depth Potential Messages}
The message from the depth variable to the depth ray potential is irrelevant since the only factor connected to the variable is the potential itself.

\subsection{Ray Depth Potential to Occupancy Variable Messages}
Similar to the depth variable messages, we marginalise out all the variables except the node in question
\begin{equation}
    \mu_{\psi_r \rightarrow o_{i}^r}\left(o_{i}^r=1\right)= 
	\sum_{d_r} \sum_{ o_j^r \atop j \neq i}  \mu_{d_r\to\psi_r}(d_r) \psi_r\left(\mathbf{o}_r, \D_r\right) \prod_{j=1}^{N_r} \mu_{o_j^r \to \psi_r}\left(o_{j}^r\right).
\end{equation}
After dropping the ray indices, we have
\begin{equation}
    \mu_{\psi \rightarrow o_{i}}\left(o_{i}=1\right)= \underbrace{\sum_{d=d_{1}}^{d_{N}} \mu(d)}_{\text{evaluates to 1}} \sum_{o_{1}} \cdots \sum_{o_{i-1}} \sum_{o_{i+1}} \ldots \sum_{o_{N}} \psi\left(o_{1}, \ldots, o_{i}=1, \ldots, o_{N}, d\right) \prod_{j=1 \atop j \neq i}^{N} \mu\left(o_{j}\right),
\end{equation}
where we use the shorthand \(\mu(d) = \mu_{d_r \to \psi_r}(d_r)\), and \(\mu(o) = \mu_{o_j^r \to \psi_r}(o_{j}^r)\). Note that as mentioned above, \(\mu(d)\) sends a uniform message to the potential since it has no other factor connected to it. Thus the outermost summation evaluates to 1.
\begin{align} \mu_{\psi \rightarrow o_{i}}\left(o_{i}=1\right)
     & = \sum_{o_{1}} \cdots \sum_{o_{i-1}} \sum_{o_{i+1}} \ldots \sum_{o_{N}} \psi\left(o_{1}, \ldots, o_{i}=1, \ldots, o_{N}, d_{j}\right) \prod_{j=1 \atop j \neq i}^{N} \mu\left(o_{j}\right)
\end{align}
We intend to simplify this expression in a similar manner to the previous derivation. Breaking apart into two terms, we have
\begin{align}
    \mu_{\psi \rightarrow o_{i}}\left(o_{i}=1\right)
     & = \mu\left(o_{1}=1\right)  \bigg[  \sum_{o_{2}} \cdots \sum_{o_{i-1}} \sum_{o_{i+1}} \ldots \sum_{o_{N}} \overbrace{ \psi\left(o_{1}=1, \ldots, o_{i}=1, \ldots, o_{N}, d\right) }^{\text{evaluates to }\nu(d_1) }\prod_{j=2 \atop j \neq i}^{N} \mu\left(o_{j}\right) \bigg]  \nonumber \\
     & + \mu\left(o_{1}=0\right)\bigg[  \sum_{o_{2}} \cdots \sum_{o_{i-1}} \sum_{o_{i+1}} \ldots \sum_{o_{N}} \psi\left(o_{1}=0, \ldots, o_{i}=1, \ldots, o_{N}, d\right) \prod_{j=2 \atop j \neq i}^{N} \mu\left(o_{j}\right) \bigg]
\end{align}
Similar to the previous strategy, we can keep breaking it up till \(o_{i-1}\). At that point we have
\begin{align}
    \mu_{\psi \rightarrow o_{i}}\left(o_{i}=1\right)
     & = \sum_j^{i-1} \mu(o_j=1) \nu{(d_j)} \prod_{k<j} \mu(o_k=0) \nonumber                                                                                                                                                                                                       \\
     & + \prod_{k<i} \mu(o_k=0) \bigg[  \sum_{o_{i+1}} \ldots \sum_{o_{N}} \underbrace{\psi\left(o_{1}=0,o_{2}=0, \ldots, o_{i-1}=0, o_{i}=1, \ldots, o_{N}, d\right)}_{\text{evaluates to}~ \nu(d_i) } \prod_{j=i+1 }^{N} \mu\left(o_{j}\right) \bigg]. \label{eq:occ_msg_common}
\end{align}
Thus, we have
\begin{equation}
    \mu_{\psi \rightarrow o_{i}}\left(o_{i}=1\right)=\sum_{j=1}^{i-1} \mu\left(o_{j}=1\right) \nu(d_j) \prod_{k<j} \mu\left(o_{k}=0\right)    +    \nu(d_i) \prod_{k<i} \mu\left(o_{k}=0\right)
\end{equation}

For the negative case we get to the same point as Eq.~\ref{eq:occ_msg_common}, except with \(o_i=0\)
\begin{align}
    \mu_{\psi \rightarrow o_{i}}\left(o_{i}=0\right)
     & = \sum_j^{i-1} \mu(o_j=1) \nu{(d_j)} \prod_{k<j} \mu(o_k=0) \nonumber                                                                                                                               \\
     & + \prod_{k<i} \mu(o_k=0) \bigg[  \sum_{o_{i+1}} \ldots \sum_{o_{N}} \psi\left(o_{1}=0,o_{2}=0, \ldots, o_{i-1}=0, o_{i}=0, \ldots, o_{N}, d\right) \prod_{j=i+1 }^{N} \mu\left(o_{j}\right) \bigg].
\end{align}
Observing that starting from \(o_{i+1}\) it is the same form of expansion, we then simplify and get
\begin{equation}
    \mu_{\psi \rightarrow o_{i}}\left(o_{i}=0\right)=\sum_{j=1}^{i-1} \mu\left(o_{j}=1\right) \nu(d_j) \prod_{k<j} \mu\left(o_{k}=0\right)    +  \sum_{j=i+1}^{N}  \mu\left(o_{j}=1\right) \nu(d_j) \prod_{k<j \atop k \neq i} \mu\left(o_{k}=0\right),
\end{equation}
Which for convenience can also be written as
\begin{equation}
    \mu_{\psi \rightarrow o_{i}}\left(o_{i}=0\right)=\sum_{j=1}^{i-1} \mu\left(o_{j}=1\right) \nu(d_j) \prod_{k<j} \mu\left(o_{k}=0\right)    +  \frac{1}{\mu(o_i=0)} \sum_{j=i+1}^{N}  \mu\left(o_{j}=1\right) \nu(d_j) \prod_{k<j} \mu\left(o_{k}=0\right).
\end{equation}

\subsection{Occupancy Variable to Ray Depth Potential Messages}
Since other rays can (and often do) pass through the same occupancy variable node, the outgoing message \( \mu_{o_{i}^r \rightarrow \psi_r } \) is computed as per Eq.~\ref{eq:node_to_factor}.